\documentclass[smallcondensed]{svjour3}     
\smartqed  

\usepackage{booktabs}
\usepackage{graphicx}
\usepackage{amsfonts} 
\usepackage{amssymb} 
\usepackage{xspace} 
\usepackage{xfrac}
\usepackage{xcolor}
\usepackage{url}
\usepackage{makeidx}
\usepackage{amsmath}
\usepackage{footnote}
\usepackage[misc]{ifsym}
\usepackage[vlined, ruled, boxed]{algorithm2e}

\usepackage{multirow}
\usepackage{subfigure}

\graphicspath{{DS19FinalSubmission/Figures/}}
 
\begin{document} 
 
 \title{Beyond Average Performance : exploring regions of deviating performance for black box classification models}

\subtitle{}

\titlerunning{Explaining the performance of classification models}        

\author{Luis~Torgo \and
        Paulo~J.~Azevedo \and
        Ines~Areosa
}

\authorrunning{L. Torgo et al.} 

\institute{Lu\'{i}s Torgo \at
         Dalhousie University, Halifax, Canada\\
         LIAAD-INESC TEC, Porto, Portugal\\
         University of Porto, Porto, Portugal \\
         \email{ltorgo@dal.ca}
         \and
         Paulo J. Azevedo \at
         Department of Informatics, University of Minho, Braga, Portugal\\
         HASLab, INESC Tec, Porto, Portugal \\
         \email{pja@di.uminho.pt}
         \and
         Ines Areosa \at
         LIAAD-INESC TEC, Porto, Portugal\\
         \email{inesareosa@gmail.com}
}

\date{Received: date / Accepted: date}






\maketitle





%
%

\begin{abstract}
Machine learning models are becoming increasingly popular in different types of settings. This is mainly caused by their ability to achieve a level of predictive performance that is hard to match by human experts in this new era of big data. With this usage growth   comes an increase of the requirements for accountability and understanding of the  models' predictions. However, the degree of sophistication of the most successful models (e.g. ensembles, deep learning) is becoming a large obstacle to this endeavour as these models are essentially black boxes. In this paper we describe two general approaches that can be used to provide interpretable descriptions of the expected performance of any black box classification model. These approaches are of high practical relevance as they provide means to uncover and describe in an interpretable way situations where the models are expected to have a performance that deviates significantly from  their average behaviour. This may be of critical relevance for applications where costly decisions are driven by the predictions of the models, as it can be used to warn end users against the usage of the models in some specific cases.

\end{abstract}

\section{Introduction}\label{sec:intro}

 Organisations are collecting large amounts of data on their activities leading to an increase in the use of Machine Learning (ML) models to automate the extraction of valuable knowledge from these data. At the same time the degree of sophistication of the available models has also been increasing steadily. Approaches like Ensembles (e.g.\cite{Zhou2012}) and Deep Learning (e.g. \cite{Zhang2018}) are among some of the most successful approaches to predictive analytics. In effect,  being able to predict the future is of key importance in many contexts and these methods have proved their value in a wide range of application domains. However, as the applicability of these methods increases, so does the awareness of the society and the visibility of some failures (e.g. \cite{Berk2018}) that have been publicly scrutinised. As a result of these social pressures, being able to understand and justify the predictions of these models has become a key goal of the research community. The main challenge of this task results from the complexity of the most successful predictive models, usually considered black boxes to illustrate the difficulty of interpreting them. This paper addresses one of the aspects of this interpretability quest: - describe, in an understandable way, situations where these black box predictive models fail or exhibit unexpected predictive performance. Finding and describing these areas of performance that deviate from the average behaviour of the models can be of utmost importance as they allow decision makers to avoid using the models on those circumstances, thus not incurring in the associated social and/or economic consequences of their erroneous predictions.

The study of the predictive performance of ML models is a well established research topic (e.g. \cite{Prati2011}). Most studies focus on characterising the global, average performance of the models through the proposal of concrete evaluation metrics (e.g. \cite{Flach2019}), or on methods to obtain reliable estimates of these metrics (e.g. \cite{TSestimaMeth}). In this paper we concentrate on drilling down this global performance analysis by focusing on finding regions of the predictors' space where the performance of the models significantly deviates from their global performance. Our methods provide the end-user with interpretable descriptions  of these areas of the input space. Note that these deviations from the average, global, performance can be in both directions. They can either be significantly worse or better than the average. We propose two types of descriptions of these deviations: (i) a univariate visual  analysis where we relate the domain of any input predictor with the expected error of the models; and (ii) a multivariate analysis where we find, and describe in an interpretable way, regions of the input space where the performance of the models is significantly different from their average performance. The former allows the user to understand how different values of a certain predictor will impact the models' performance, while the latter discovers areas of the input space where the performance of the models is significantly worse (or better) than the global behaviour. 

The two tools we propose allow the end user to explore in an interpretable way the error profile of any classification model. These tools provide a level of detail of their predictive performance that can be used to anticipate situations where using the models to drive decisions may be too risky. In this context, these tools empower end users with the ability to anticipate the risks of using machine learning models to drive decisions, by providing interpretable accountability summaries of their expected performance. In the paper we describe a series of concrete illustrative examples of how these tools can help in identifying these high-risk situations.

The paper is organised as follows. In the next section we describe existing work that is related to our proposals. Section~\ref{sec:method} describes our two proposed methods, while on Section~\ref{sec:exps} we provide concrete examples of applying these methods to better understand the performance profile of different classification methods. Finally, we present the conclusions of our work on Section~\ref{sec:conclusion}.

\section{Related Work}\label{sec:related_work}

Existing work hinges on distinct aspects of explainability, with most methods aiming to explain the possible cause and effect phenomenons that are responsible for the concrete values predicted by a model. However, assessing and understanding the performance  of a black-box model is also fundamental to enhance accountability, while improving the knowledge of the machine.

For both cases, the explanations can be provided through a multitude of means, such as visual aids or textual descriptions. While visualization is more convenient for compiling complex ideas, textual explanations have the advantage of transmitting information in human-like manner and of being straightforward. For the latter, Association Rules \cite{AR} are frequently used as highly interpretable means of correlating different facts about observations \cite{LORE,Lormika,Lime}.

Several tools for evaluating classification models can be found in the literature, ranging from scalar to graphic methods. These include, for instance, the largely used metrics of accuracy (or the complement error rate), precision, recall and the F1-score \cite{fscore:rijsbergen} that can be calculated recurring to a confusion matrix. Other approaches try to provide a different perspective on the analysis of the model, informing about the changes in the performance for different operating conditions, as are examples ROC curves \cite{Collinson:ROC}, the Area Under the Curve (AUC) \cite{hanley:AUC} and Cost Curves \cite{drummond:CC}.

 
The shortcoming of using  single scalar measures is that these cannot capture the full complexity of the performance profile of a model,  neither help in differentiating between two classifiers across particular types of test cases. Although ROC curves and Cost Curves contribute to overcome this issue, these methods still fail in analysing which are the circumstances influencing the performance of a machine, i.e. they fail in associating the values of the predictors with these different performance levels thus not providing interpretable descriptions of the concrete regions of the predictors space that cause  this behaviour.

In prior work, we have presented the Error Dependence Plot (EDP) \cite{EDPjournal}, which is a visual tool that helps identifying under which conditions of the predictor variables the performance of a certain black box regression model will deviate significantly from the expected overall performance. In this paper we  extend these methods to classification models.

Several proposals are described in the literature that make use of rules to explain predictions. One of the first was \textit{anchors} \cite{anchors}, where individual predictions are explained with decision rules that are generated through perturbations that evaluate the local changes of behaviour of the model.
LORE proposed in \cite{LORE} is an agnostic method (the algorithm can be used in any type of machine learning model) to derive interpretable and faithful explanations. It uses a genetic algorithm to derive decision rules that explain the reason for the prediction and a set of counterfactuals to explain which changes are needed in the instance features state to flip the prediction. The method only applies to binary classification problems.
The Lormika method described in \cite{Lormika} also uses association rules as the local model to explain an instance. These k-optimal association rules are derived from the instance neighbourhood. The authors argue that these k-optimal rules are the best rules to explain the prediction since they follow an optimised algorithm on a specific interest measure e.g. lift, strength, leverage. Lormika also generates counterfactual rules to suggest which potential changes in the instance features state lead to different outputs on the model prediction. All these approaches once again focus on explaining the predicted values, not the expected error which is the goal of the current paper.

In this paper we will recover distribution rules \cite{DR06}, a type of association rules that describe subgroups with a deviating numerical property. The derived rules (and subgroups) exhibit a distribution on this property that significantly distances from a given reference distribution (typically the global numeric property distribution). In this work, we adapt the proposal in \cite{DR06} to categorical distributions so that one can deal with a performance representation of classification models. In our case, this representation is the confusion matrix of a model in a given test set.  

In light of previous investigation, we found that performance tools that assess classification models solely address the error or the error tolerance in relation to the target value, never establishing a relationship between these errors and the values of the predictors.  

Henelius et al.~\cite{Henelius2014} studies hard classifiers and try to find groups of attributes whose interaction affect the predictive performance of a given model. They propose a GoldenEye algorithm that makes use of randomisation of data and a fidelity measure to identify optimal sets of attributes. 

Duivesteijn et al.~\cite{SCAPE} (SCaPE) appear as the closest proposal to ours. The authors make use of Exceptional Model Mining (EMM) \cite{EMM} for finding subgroups for which a soft classifier performs poorly or exceptionally in relation to a given ground truth. In SCaPE, EMM is used to detect abnormal interaction between multiple properties.
Used datasets in SCaPE (test sets) have the effective class
value and the classification score of the model included as two additional attributes.
The interaction between these two properties (effective class and classification score) on each test case is used to compute a rank based measure. When compared to the complete test set performance, this measure enables an ordering on subgroups (using an AUC rank like measure).
This measure (Average Ranking Loss, ARL) computes a penalty for each positive case in the dataset. This penalty corresponds to the number of negative cases that have higher score than the positive case. To identify the subgroups that represent the most extreme situations, the difference between the ARL of the whole test set and the ARL of the cases covered by the subgroup is calculated. Hence, it requires two runs of SCaPe to obtain the set of subgroups representing both extreme situations.
SCaPE application requires a soft classifier where a score is always derived for each prediction. It also implies that this proposal is only applicable to binary classification problems. These two features makes comparison with our proposal unfeasible.
The specific rates of false positives/negatives are not displayed within SCaPE output. Also, it is not possible to follow how subgroup specialisation evolves since no rule pruning control is used.
We also notice that the subgroups in the case study described in \cite{SCAPE} are only composed of one condition which yields a rather simplified approach.

\section{Interpretable Analysis of Classification  Performance}\label{sec:method}

Our main goal is to design and formulate accountability methods that focus on describing regions of the predictors space where a certain unusual error behaviour occurs. Specifically, we will provide two solutions to this: (i) a univariate proposal that describes how changes across the domain of a predictor variable affect the expected model performance; and (ii) a multivariate method that provides interpretable descriptions of areas of the predictors space where the performance of the models is significantly different from their average behaviour captured by standard scalar evaluation metrics.

\subsection{Univariate Methods}\label{sec:EDPs}

Regression EDPs \cite{EDPjournal} are visual tools that represent the distribution of the expected error of a regression model on the Y-axis, against the values of a predictor variable in the X-axis. In theory, the procedure used to obtain these plots would consist of obtaining reliable estimates of the prediction error for each value of the predictor.  However, calculating the estimated error for each  possible value of a numerical predictor is challenging, since each value might not repeat often or even appear in the available data, particularly when dealing with smaller datasets. Hence, the practical approach that was taken was to compartmentalise  the domain of the numeric predictors  into relevant bins, and then estimate the error of the models for each of these bins. To make sure these estimates (and thus EDP's) are reliable  a Cross Validation procedure was used, making sure each of the  cases in the available dataset is used in a test set once. Using the estimates of the error of the model for each case we can obtain the distribution of this error for each of the bins. 

The discretization of the domain of the numeric predictors can be driven by specific user/domain requirements or, in the absence of these, as suggested in the original paper,  the domain can be divided into 5 bins, according to the quantiles of the values: $[0, 10\%]$ (extremely low values), $[10\%, 35\%]$ (low values), $[35\%, 65\%]$ (central values), $[65\%, 90\%]$ (high values) and $[90\%, 100\%]$ (extremely high values). For nominal predictors this division is not necessary since the variables are discrete, and each value will already act as a bin. 

The previously developed regression EDPs show the estimated error distribution of each bin through a boxplot, since  regression errors are continuous. This is not the case for classification tasks that are addressed in the current paper. Due to the nature of  categorical prediction errors,  EDPs for  classification  will have to focus on analysing  the estimated distribution of misclassifications for each bin of the predictors. Contrary to regression error, classification errors have a finite domain and are typically described by a confusion matrix with $c\times c$ dimension, where $c$ is the number of classes of the problem. Our proposal uses again a Cross Validation procedure to estimate in a reliable way the numbers in the confusion matrix. These numbers are then partitioned according to the bins of the predictor for which a classification EDP is required. The other necessary modification to the original EDP's concerns the procedure used to show this classification error distribution. Given the nature of the information we have opted for displaying  these confusion matrices as  stacked bar plots where each bar,  representing one entry of the matrix,  shows the percentage of the results with that particular misclassification error. 

\begin{figure}[h!]
\centering
\subfigure[EDP]{\includegraphics[width=0.99 \textwidth]{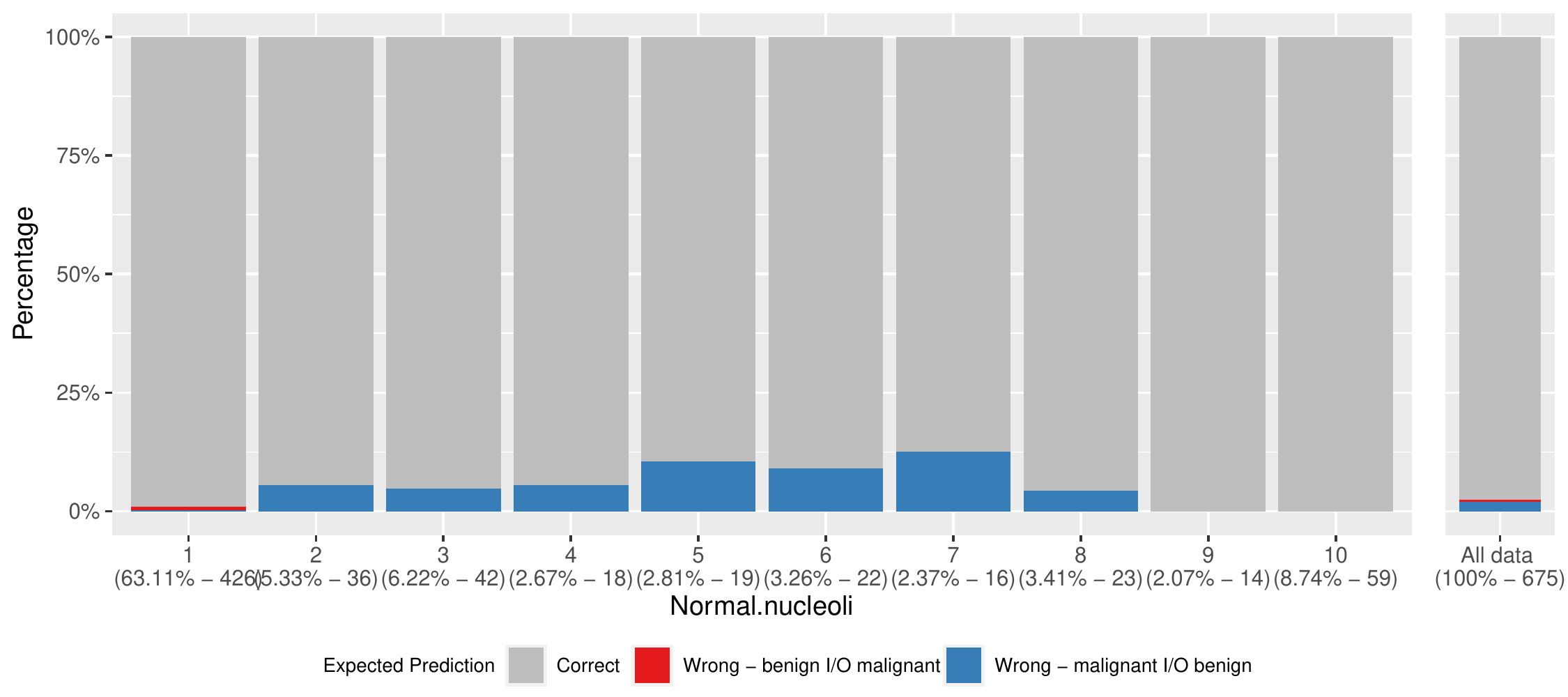}}

\subfigure[Error-Zoom  EDP]{\includegraphics[width=0.99 \textwidth]{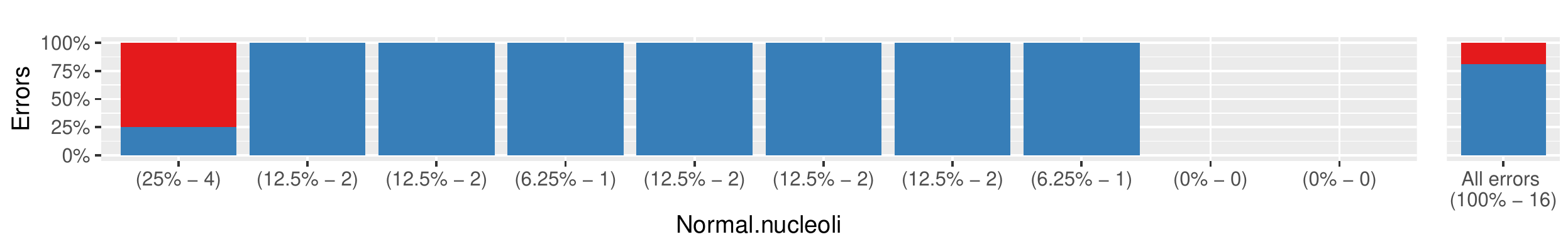}}
\caption{Error Dependence Plot for feature \textit{normal.nucleoli} from dataset \textit{BreastCancer} trained with Naive Bayes  (c.f. Tables \ref{tab:datasets} and \ref{tab:modelsbench}).}
\label{fig:edp_breast}
\end{figure}

Figure~\ref{fig:edp_breast} a) depicts an example of a classification EDP of a Naive Bayes (NB) model for the numerical predictor \textit{normal.nucleoli} of dataset \textit{BreastCancer} (c.f. Tables \ref{tab:datasets} and \ref{tab:modelsbench}). Below each bin of \textit{normal.nucleoli} we present the information on the number of training cases in the bin and the respective percentage of the full data set. Moreover, for comparison,  EDPs  visualize the  error  distribution  over the entire  data  set  on  the  right  side  of the  plot, i.e. the expected global performance of the model. The plot dissects the performance of the model across the range of the predictor, helping to understand, for instance,  that the model is expected to underperform for central values (\textit{normal.nucleoli}~=~[5-7]), where there is an higher percentage of errors, and to overperform for higher values (\textit{normal.nucleoli}~=~[9-10]), where no errors are expected to occur. It is also interesting to note that for the cases in which only one nucleoli is found, there is a higher  expected risk of a false negative (predicting a benign tumour instead of a malign), which does not occur for any other value of that predictor. This simple example immediately highlights the relevance of this drilled down performance analysis that may lead decision makers to avoid using the predictions of a model in face of such different expected error. Knowing that if a certain test case contains a single nucleoli the NB model incurs in a high risk of saying that a tumour is benign instead of malign may be crucially important information for a decision maker. This information cannot be inferred from the global analysis of the model performance.

Since in some models the expected percentage of errors for a bin can be too small to analyse visually on the naked eye, as it is the case of  \textit{normal.nucleoli}~=~[1] (Figure~\ref{fig:edp_breast} a)),  classification EDPs provide an optional error-zooming plot for a closer look at the errors, by eliminating the accurate predictions from the visualisation. An example can be found in Figure~\ref{fig:edp_breast} b), in which each bar represents the errors found in the bin of \textit{normal.nucleoli} shown in Figure~\ref{fig:edp_breast} a). The percentage below each bin of this zoomed plot refers to the percentage of total errors of the model. For instance, from the 18 (100\%) errors of NB (rightmost stacked bar on Figure~\ref{fig:edp_breast} b) ), 25\% (4) occur when \textit{normal.nucleoli}~=~[1], and 3 out of 4 of  these errors (75\%) are false positives (predicting a benign tumour when it is malign). We can also observe from this graph that all false positive errors, which in this application domain are the most serious ones, occur when \textit{normal.nucleoli}~=~[1]. This is a crucial piece of information concerning the predictive performance of this NB model, that can not be inferred from the global analysis of its performance. If such model is being used for supporting medical decisions, this graph would provide a clear red flag when a patient with \textit{normal.nucleoli}~=~[1] appears.



In summary, the proposed method is based on two main steps: (i) obtain a reliable estimate of the predictive performance of a model; and (ii) drilling down this estimated predictive performance across the bins of any predictor variable, representing the outcome as stacked bar plots.

For the first of these steps any estimation methodology could be used. In our case we have used $k$-fold Cross Validation (CV). More specifically, we have used CV to obtain the prediction of any classification model for all available cases. By definition of CV, any case of a dataset is part of one of the $k$ folds that are used as a test set in the CV procedure. This means that for each case, we will be able to obtain the prediction of the model in a reliable way (i.e. when the model has not used that case in its training set). Using these predictions for all cases in the data set, obtained in the CV process, it is possible to derive confusion matrices that summarise the model performance on the data set. Algorithm~\ref{alg:cv} describes the procedure used to obtain such confusion matrix through CV.
 
\begin{algorithm}[]
    \SetKwFunction{train}{Train}
    \SetKwFunction{pred}{Predict}
    \SetKwFunction{perm}{Permute}
    \SetKwFunction{part}{Partition}
    \SetKwFunction{cm}{ConfMtrx}
    \SetKwInOut{Input}{input}
    \SetKwInOut{Output}{output}
    \SetKwInOut{s}{s}

    \Input{data set $\mathcal{D}$}
    \Input{algorithm $\mathcal{A}$}
    \Input{nr. folds $k$}
    \Output{cases and their predictions $\hat{E}$}

    \BlankLine
    
        $\mathcal{D}' \leftarrow$ \perm{$\mathcal{D}$} \tcp{randomly permute the data}
        $P \leftarrow$ \part{$\mathcal{D}'$, $k$}  \tcp{create $k$ equal-size partitions}
        $\hat{E} \leftarrow \{\}$

        \ForEach{$p$ in $P$} {
            $M \leftarrow $ \train{$\mathcal{A}$, $\mathcal{D}' \setminus\mathcal{D}'_p$}   \tcp{train $\mathcal{A}$ on all but the partition $p$ cases}
            $\hat{e}_p \leftarrow \{$ $\langle\mathbf{x}, y, \hat{y}\rangle$ $ |\ \langle\mathbf{x}, y\rangle\, \in\mathcal{D}'_p\ \land\ \hat{y} = $ \pred{M,$\mathbf{x}$} $\}$ \tcp{test cases in $p$ and respective predictions}
            $\hat{E} \leftarrow \hat{E} \cup \hat{e}_p$
        }
    
    \Return $\hat{E}$ \tcp{Return the cases and respective predictions of the model}
\caption{Obtaining Cross validation Predictions of a Classifier.}
\label{alg:cv}
\end{algorithm}

For the second step we need to obtain the error profile of the models for each bin of a predictor variable. This profile can be obtained by using the information returned by Algorithm~\ref{alg:cv}. Specifically, for a bin $b$ of a predictor we just need to obtain the subset of cases in the set $\hat{E}$ returned by that algorithm, that have a value of the predictor inside that bin. With this subset we can obtain the number of correct predictions and also the errors of those cases, which is then represented by a stacked bar. This procedure is formalised in Algorithm~\ref{alg:EDP}.

\begin{algorithm}[h!]
    \SetKwFunction{bin}{FindBin}
    \SetKwFunction{getbins}{DefineBins}
    \SetKwFunction{cat}{Categories}
    \SetKwFunction{sbp}{DrawStackedBar}
    \SetKwFunction{com}{ConfusionMtrx}
    \SetKwInOut{Input}{input}
    \SetKwInOut{Output}{output}

    \Input{data set $\mathcal{D}$}
    \Input{predictions $\hat{E}$ for the cases in $\mathcal{D}$}
    \Input{bins $B$ of the predictor $X^k$}

    \BlankLine

        \If{$B$ is empty}{
        \If{$X^k$ is numeric}{
            $B \leftarrow$ \getbins{$X^k$}   \tcp{get the bins of  $X^k$ using quantiles or user-defined ranges}
        }
        \If{$X^k$ is nominal}{
         $B \leftarrow$ \cat{$X^k$}   \tcp{get the bins of  $X^k$ using categories}
        }
        }

        \ForEach{$b$ in $B$} {
            $E_b \leftarrow \{\}$
        }

        \ForEach{$\langle\mathbf{x}_i,y_i\rangle$ in $\mathcal{D}$} {
            $b \leftarrow $ \bin{$x^k_i$, B} \tcp{get the bin of the value of $X^k$}
            $E_b \leftarrow E_b \cup \{$ $\langle y_i, \hat{y}_i\rangle$ $ |\ \langle\mathbf{x}_i, y_i, \hat{y}_i\rangle\, \in\hat{E}$ $\}$ \tcp{true and predicted values for this case}
        }

        \ForEach{$b$ in $B$} {
            $CM_b \leftarrow $ \com{$E_b$} \tcp{Calculate the confusion matrix aggregating the correct prediction numbers into a single score}
            \sbp{$E_b$}
        }

\caption{Obtaining the classification EDP of a predictor.}
\label{alg:EDP}
\end{algorithm}

\subsection{Multivariate Methods}\label{sec:ARs}

\vspace{2mm}

Distribution rules \cite{DR06} are a form of association rules that discover subgroups with distinguished properties of interest. The idea is to adapt an association rules algorithm to derive rules that find subgroups that have a deviating distribution  in a predefined numeric property of interest. These distributions are deviating compared to a prior distribution. This prior distribution is typically the distribution of the whole population. Measuring deviation of a distribution in relation to a predefined one (prior) is implemented using a goodness of fit statistical test. The original proposal used the two sample Kolgomorov-Smirnov continuous distribution significance test (KS-test).
The following is an example of a distribution rule (from the \textit{wages} dataset, c.f. Table \ref{tab:datasets}):

\vspace{2mm}
\begin{verbatim}

  Ant Sup=0.118  pvalue=0.0085  Mean=10.982   St.Dev=6.333
  WAGE <-- education=]12.5 : 15.5] &  SOUTH=0 &  RACE=3.
\end{verbatim}
\vspace{2mm}

\noindent and can be read as: "the subgroup composed by white people (\textit{race}=3) with 13 to 16 years of education not from the south (\textit{south}=0) has a wage distribution that deviates significantly from the distribution of the whole population". The  wage distribution of the whole is not shown for space convenience. $\tt Ant\; Sup$ represents subgroup frequency, $\tt pvalue$ refers to the KS-test, $\tt Mean$ and $\tt St.Dev$ are the property of interest mean and standard deviation for the subpopulation covering this subgroup.

In this paper we revisit distribution rules to assist in drilling down the error performance analysis of classification models. We take as input a categorical distribution formed with bins that represent cells in the model's confusion matrix estimated through the same process used in classification EDP's (see Algorithm~\ref{alg:cv}). The diagonal (correct predictions) is taken as a single bin in this categorical distribution (as we have done for EDP's). To enable the use of these new type of distributions, a new module for categorical distribution rules was implemented in CAREN \cite{caren} where distance between distributions is measured using a $\chi^2$ goodness of fit test.
The same algorithm used to find the continuous distribution rules is then applied, but this time driven by this different statistical test that is required for comparing categorical distributions.

The data set used to find the distribution rules that characterise  the performance of a model  is created using the original predictor variables plus a categorical property of interest that in this case represents prediction performance, i.e. the cells in the confusion matrix (CM). The coding for these cells follows the pattern  effective/predicted. Thus, code 12 means an instance of class 1 where the model predicted class 2. Code 0 represents a hit (the diagonal of the CM).

Consider an example from a binary classification task (\textit{adult} dataset, Table \ref{tab:datasets}) where class 1 is the  class "income $\leq$ 50K"   and class 2 is "income $>$ 50K". Suppose a certain classification model is applied to the full data set resulting in a confusion matrix whose distributions is : 

\vspace{2mm}
\begin{verbatim}
CM={ 0/0.854,12/0.044,21/0.102 }
\end{verbatim}
\vspace{2mm}

\noindent this means that in the full data set the classification model was accurate in 85.4\% of the cases, with 4.4\% being cases of class 1 wrongly predicted as class 2, and the remaining 10.2\% being class 2 cases classified as 1.

The following  categorical distribution rule describes, in an interpretable way, an interesting subgroup of cases from the point of view of classification performance. This subgroup is interesting because the model classified it in such a way that resulted in a confusion matrix whose categorical distribution significantly deviates from the above distribution. The rule is the following:  

\vspace{2mm}
\begin{verbatim}
Ant sup = 0.01078  pvalue = 0.0010000156478771152000  
CM={ 0/0.880,12/0.006,21/0.114 }  <--    
            education=Bachelors  & relationship=Not-in-family  & 
            occupation=Prof-specialty  &  workclass=Private
\end{verbatim}
\vspace{2mm}

\noindent The rule can be read as: "The performance of the model when applied to the subgroup formed out of the  cases having people holding a Bachelor, not having a family, having a specialised profession and working in the private sector deviates significantly from the performance exhibited by the model in the full data set. For this subgroup of  cases the ratio of false positives (noted as "21" above) is 11.4\% (10.2\% in the full  set) for class 1 and 0.6\% (4.4\% in the full  set) of false negatives for class 2. The hit rate (accuracy) is 88.8\% (85.4\% in the full  set)). 
The categorical distribution of the error for this subset of cases significantly deviates ($pvalue$ of 0.001) from the distribution in the full set according to the used statistical test. The subgroup frequency (Ant sup) is 1.078\% of the dataset.


In the original CAREN proposal~\cite{DR06}, rule pruning was implemented using subrule comparison. Pruning is achieved using the idea of rule improvement. A rule is considered whenever a comparison with its subrules yields an interest measure improvement. The $pvalue$ was used as the interest measure. Thus, a rule is derived whenever the $pvalue$ is lower than all its subrules $pvalues$. We adopt the same strategy and for instance, the rule above is an improvement to the following:

\vspace{2mm}
\begin{verbatim}
Ant sup = 0.01772  pvalue = 0.0010378215125190922000  
CM={ 0/0.887,12/0.007,21/0.106 }    <--    
        education=Bachelors  &  relationship=Not-in-family  & 
        occupation=Prof-specialty    
\end{verbatim}
\vspace{2mm}
    
It is interesting to notice in this case that the false negative rate for class 2 increased with rule specialisation.


\vspace{2mm}
In summary, the procedure to obtain these categorical distribution rules describing  regions of the predictors space where deviating error behaviour occurs,
consists of the following main steps:
\begin{enumerate}
\item Obtain prediction values for each test set case through cross validation using a specific classification algorithm on the original dataset

\item derive a new dataset with an additional column which will be the property of interest for the distribution rules. This new attribute is the composition of the effective class value with the predicted one.

\item discretize numerical attributes according to the given set of bins (typically quantiles).

\item derive categorical distribution rules in this new version of the dataset using the new property of interest.

\end{enumerate}

\begin{algorithm}[]
    \SetKwFunction{train}{Train}
    \SetKwFunction{pred}{Predict}
    \SetKwFunction{perm}{Permute}
    \SetKwFunction{part}{Partition}
    \SetKwFunction{cm}{ConfMtrx}
    \SetKwInOut{Input}{input}
    \SetKwInOut{Output}{output}
    \SetKwInOut{s}{s}

    \Input{data set $\mathcal{D}$}
    \Input{algorithm $\mathcal{A}$}
    \Input{nr. folds $k$}
    \Input{set of bins $Bs$ for each of the predictors}
    \Input{minsup value $ms$}
    \Output{set of rules $SDRs$}

    \BlankLine
    
        $\mathcal{D}' \leftarrow$ \perm{$\mathcal{D}$} \tcp{randomly permute the data}
        $P \leftarrow$ \part{$\mathcal{D}'$, $k$}  \tcp{create $k$ equal-size partitions}
        $\hat{E} \leftarrow \{\}$

        \ForEach{$p$ in $P$} {
            $M \leftarrow $ \train{$\mathcal{A}$, $\mathcal{D}' \setminus\mathcal{D}'_p$}   \tcp{train $\mathcal{A}$ on all but the partition $p$ cases}
            $\hat{e}_p \leftarrow \{$ $\langle\mathbf{x}, y \bigoplus \hat{y}\rangle$ $ |\ \langle\mathbf{x}, y\rangle\, \in\mathcal{D}'_p\ \land\ \hat{y} = $ \pred{M,$\mathbf{x}$} $\}$ \tcp{test cases in $p$ and concatenation between effective and predicted class}
            $\hat{E} \leftarrow \hat{E} \cup (\hat{e}_p$)
        }
    $E_{new} = Discretize(\hat{E},Bs)$ \tcp{discretize new dataset using the given bins}
    $SDRs = CatDistriRules(E_{new},ms)$ \tcp{obtain a set of distribution rules on dataset $E_{new}$ using minsup $ms$}
    \Return $SDRs$ \tcp{Return the set of distribution rules}
\caption{Extracting a set of distribution rules from a new dataset with the categorical property of interest obtained through Cross validation.}
\label{alg:cv3}
\end{algorithm}

Algorithm~\ref{alg:cv3} is similar to Algorithm~\ref{alg:cv} and formalises the four described steps. It derives a new dataset with a categorical property of interest for extracting distribution rules. Then it discretizes the numeric attributes according to the given bins. Finally  applies the modified CAREN distribution rules algorithm to this discretized dataset using the new categorical property of interest representing the model's confusion matrix.

\section{Experimental Analysis}\label{sec:exps}

This section focuses on describing a series of illustrative case studies that provide evidence of the practical usefulness of the tools we have proposed to describe and understand   the performance of a set of black-box classification models. 

Considering the high number of datasets, predictors and models in analysis, not all of the results can be shown here. Hence, the complete set of  plots and rules can be seen in the web page \url{https://ltorgo.github.io/ExplainClass/}. The same web page contains all code and CAREN version used, ensuring full reproducibility of our results and analysis.

\subsection{Material and Methods}\label{sec:ma}

We have experimented with our methods on  18 datasets from different domains, with variable size, number of predictors and of number of classes of the target variable, as described in Table~\ref{tab:datasets}. These are publicly available in \url{https://ltorgo.github.io/ExplainClass/}.




\begin{table}[t]
\centering

\begin{tabular*}{0.6\textwidth}{l  r r r }
\hline
\textit{Dataset}& \textit{\#Inst} & \textit{\#Pred} & \textit{\#Classes} \\
\hline
breast-base &$ 86 $&$10$ &$2$ \tabularnewline
contraceptive &$ 1473 $&$10$ &$3$ \tabularnewline 
lymphography &$ 148 $&$19$ &$4$ \tabularnewline
soybean &$ 268 $&$36$ &$15$ \tabularnewline
yeast3 &$ 1484 $&$9$ &$2$ \tabularnewline
BreastCancer &$ 699 $&$10$ &$2$ \tabularnewline
Glass &$ 214 $&$10$ &$6$ \tabularnewline
PimaIndiansDiabetes &$ 768 $&$9$ &$2$\tabularnewline
iris &$ 150 $&$5$ &$3$ \tabularnewline
LetterRecognition &$ 20000 $&$17$ &$26$\tabularnewline
Vehicle &$ 846 $&$19$ &$4$ \tabularnewline
Vowel &$ 990 $&$11$ &$11$ \tabularnewline
scat &$ 110 $&$19$ &$3$ \tabularnewline
HeartDisease &$ 303 $&$14$ &$2$ \tabularnewline
Wine &$ 178 $&$14$ &$3$ \tabularnewline
Adult &$ 32561 $&$15$ &$2$ \tabularnewline
Yeast &$ 1484 $&$9$ &$10$ \tabularnewline
Fertility & 100 & 11 &$2$ \tabularnewline
Wage &$ 534 $&$11$ & $-$ 
\tabularnewline
\hline \tabularnewline
\end{tabular*}

\caption[Datasets used for benchmarking]{Datasets used for benchmarking ({\small $\#Inst$: number of instances; $\#Pred$: number of predictor variables; $\#Classes$: number of classes of the target variable }).}
\label{tab:datasets}
\end{table}


Each of the datasets was modelled as a classification task using 4 distinct  learning algorithms, with characteristics described in Table~\ref{tab:modelsbench}. The diversity of models selected (Naive Bayes, Random Forest, Neural Network and Support Vector Machine) avoids the existence of model-dependent bias on our experimental observations.

\begin{table}[t]
\centering
\scriptsize
\begin{tabular*}{ \textwidth}{l @{\extracolsep{\fill}} lr}
\hline 
Learner & Parameters & R package \\ 
\hline 
NN & $size=10,decay=0.1,maxit=1000, MaxNWts = 32561$ & \textbf{nnet}~\cite{nnet} \\ 
SVM & $cost=10, gamma=0.01$  & \textbf{e1071}~\cite{e1071} \\

RF & $ntree=1000$& ~~ \textbf{randomForest}~\cite{rf} \\ 

NB & $laplace=0$ & \textbf{e1071}~\cite{e1071} \\
\hline \tabularnewline
\end{tabular*} 
\caption{Classification algorithms, parameters, and respective R packages used for the benchmarking.}
\label{tab:modelsbench}
\end{table}

The  tools we have proposed to understand the performance of the models are based on estimates of the expected prediction errors of these models. As such, to ensure the analysis of the results is reliable, a 10-fold Cross Validation process was used to obtain the prediction of the models for each case in the data sets, using the R package \textit{performanceEstimation}~\cite{torgo2014infra}. In Cross Validation, each of the 10 folds is used as testing dataset once. This means that for each case we are able to obtain a prediction of the model (learned with a separate training set), together with the true value of the target variable of the case.  Comparing these predictions to the true values we obtain reliable estimates of the error of the model. Using this procedure we obtain a full confusion matrix for all cases on each data set that will be the input to our proposed explainability methods.

The univariate methods were implemented in R \cite{rprog} using the \textit{ggplot2}~\cite{ggplot2} package, while the multivariate method uses a new CAREN implementation \cite{caren} with the proposed new categorical distribution rules module. The minimal support used for rule extraction in all datasets was 1\%.

\subsection{Univariate Methods}\label{sec:evalEDPs}

\begin{figure}[h!]
\begin{center}
  \includegraphics[width=0.99 \textwidth]{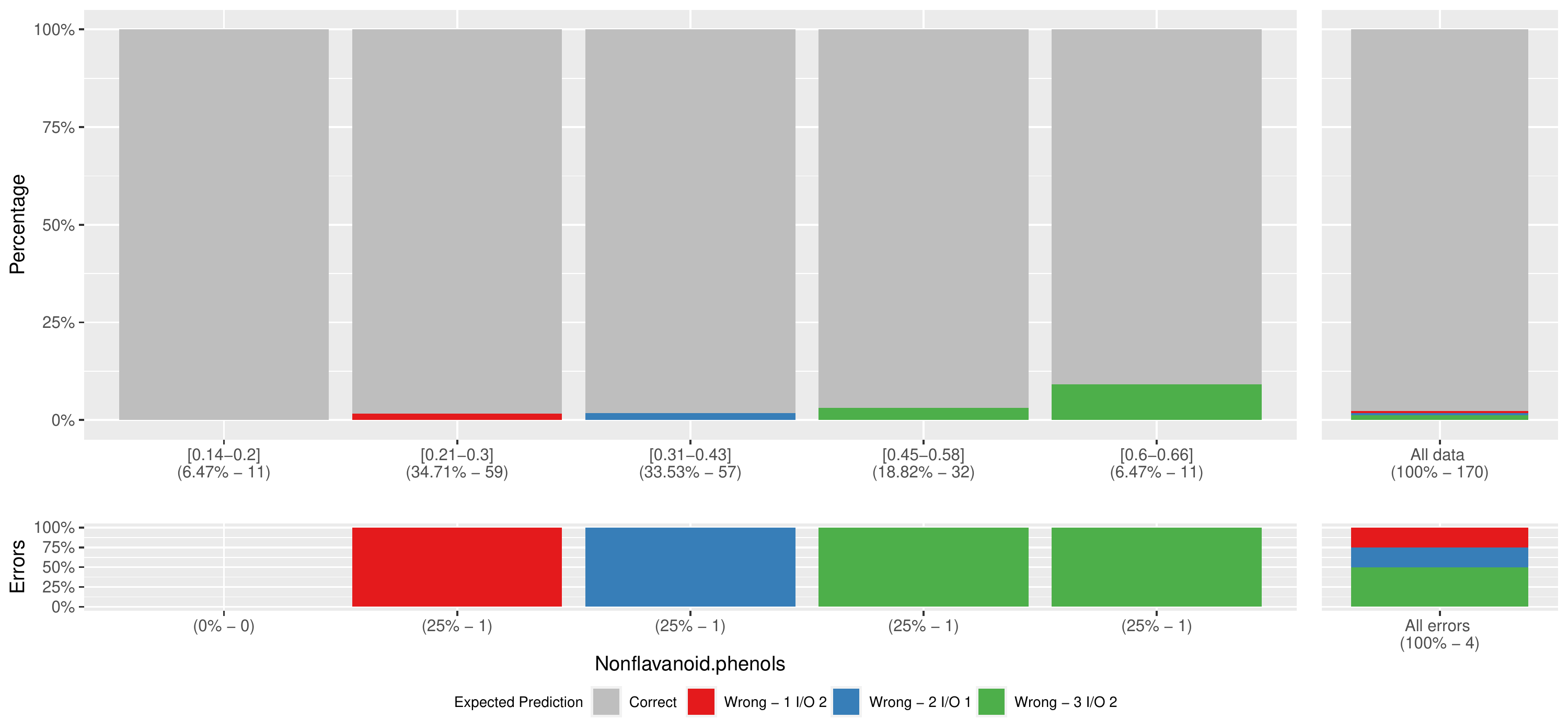}
\caption[]{Error Dependence Plot for feature \textit{Nonflavonoid.phenols} from dataset \textit{Wine} trained with Random Forest.}
\label{fig:wine}
\end{center}
\end{figure}

Figure \ref{fig:wine} shows the error distribution of a Random Forest model in the task of classifying 3 different types of wine (classes \textit{1, 2} and \textit{3}) regarding the content of non-flavonoid phenolic constituents. This EDP shows that different misclassifications occur for distinct values of this feature - for instance, a classification of class 1 instead of 2 occurs only for low values (\textit{Nonflavonoid.phenols}~=~$[0.21-0.3]$), while a wrong classification of 3 instead of 1 occurs for high and extremely high values of the predictor (\textit{Nonflavonoid.phenols} = $[0.45-0.58]$ and \textit{Nonflavonoid.phenols}~=~$[0.6-0.66]$).

\vspace{2mm}

\begin{figure}[h!]
\begin{center}
  \includegraphics[width=0.99 \textwidth]{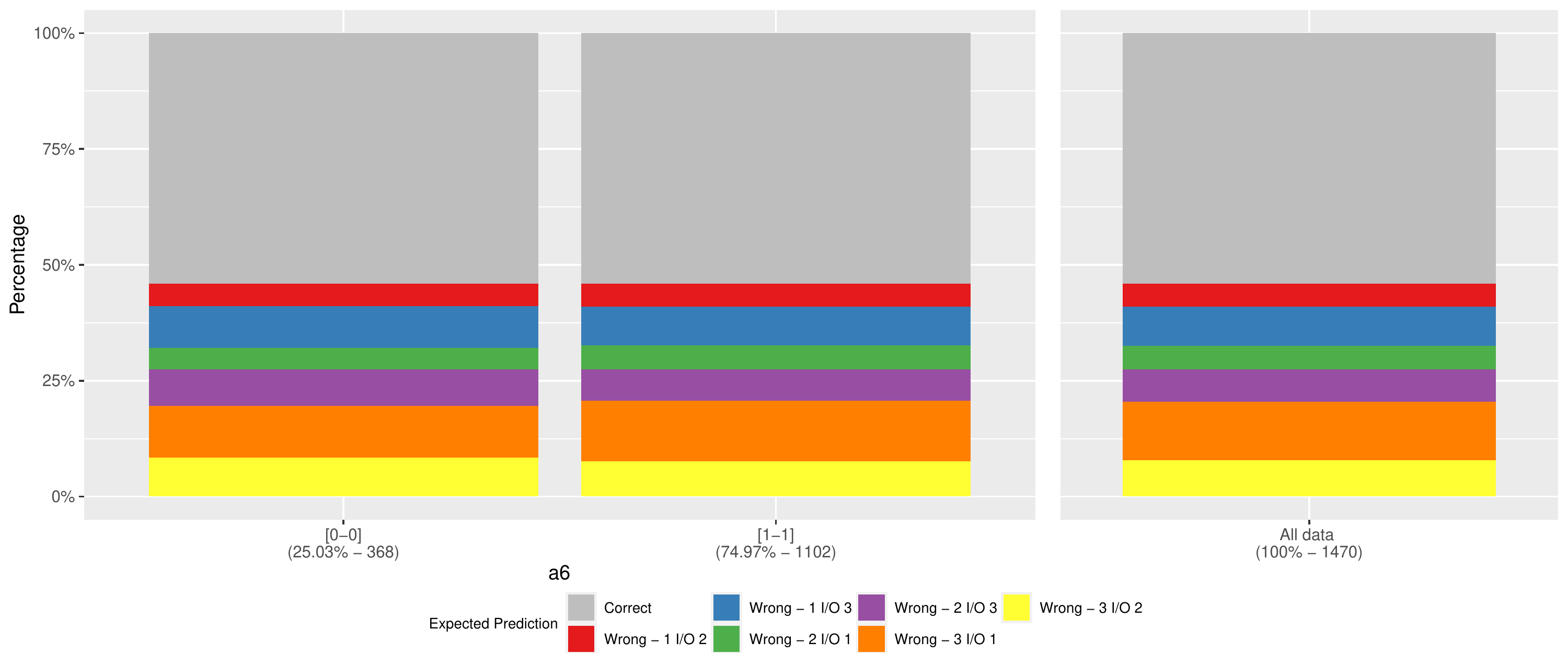}
\caption[]{Error Dependence Plot for feature \textit{a6} from dataset \textit{contraceptive-base} trained with Neural Network.}
\label{fig:contraceptive}
\end{center}
\end{figure}

EDPs are not only fit for finding distinctive patterns of error distributions, as they can also help in assuring that the performance of a model is not dependent on the value of a particular predictor variable. Figure~\ref{fig:contraceptive} illustrates one of these cases, for dataset \textit{contraceptive-base}, when modelled using a NN. This EDP helps understanding that the distribution of the expected error is very similar for each possible value of \textit{a6}, showcasing that this feature does not influence (at least directly) the performance of the NN in analysis.

\vspace{2mm}

\begin{figure}[h!]
\begin{center}
  \includegraphics[width=0.99 \textwidth, height = 0.33\textheight]{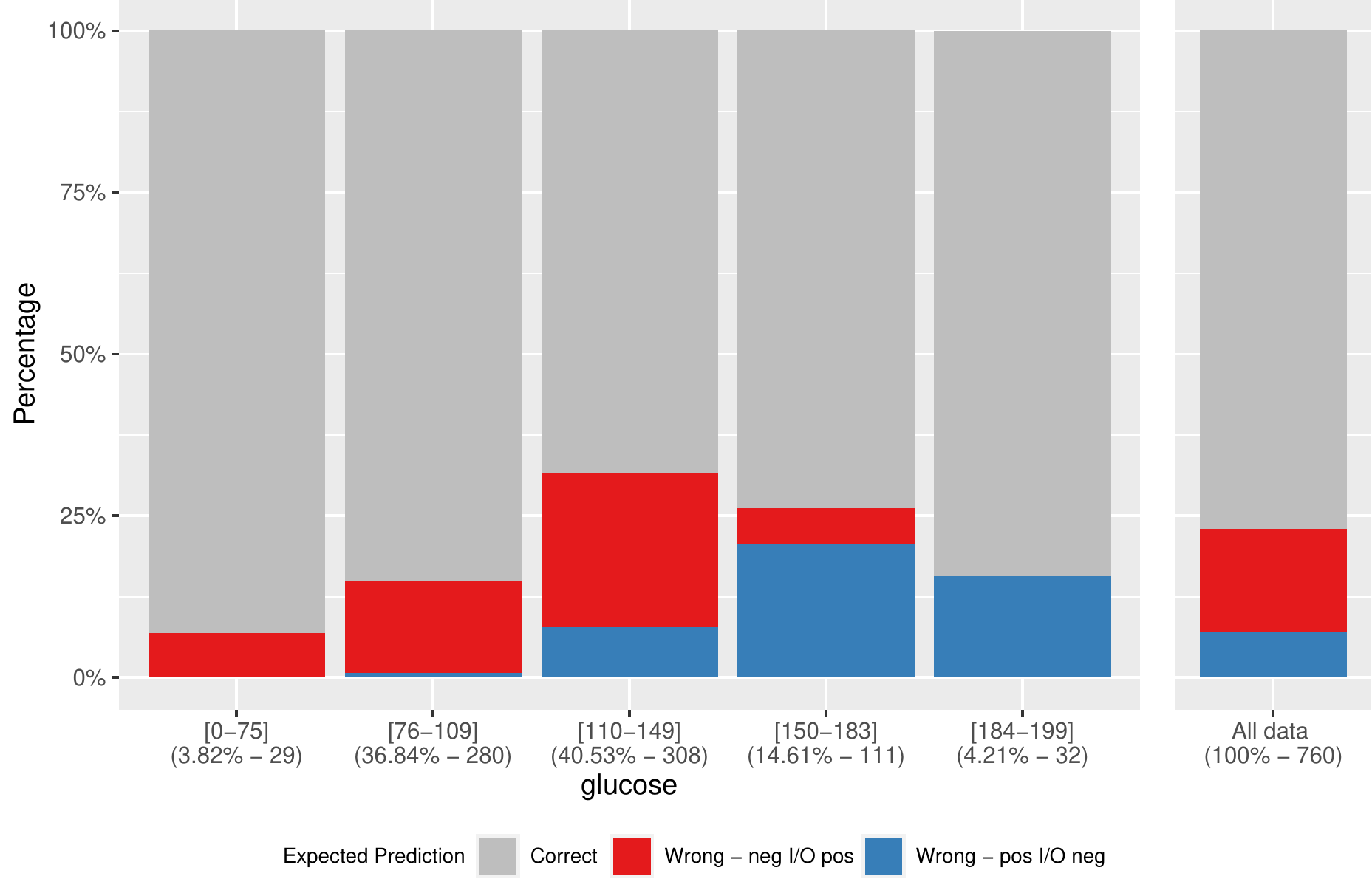}
\caption[]{Error Dependence Plot for feature \textit{glucose} from dataset \textit{PimaIndiansDiabetes} trained with Support Vector Machine.}
\label{fig:pima}
\end{center}
\end{figure}

Figure \ref{fig:pima} shows one example of the EDP of a SVM for dataset \textit{PimaIndianDiabetes}. We can observe that the performance of the model varies considerably depending on the value of \textit{glucose}.  More concretely, the SVM is expected to have a worse performance for central values (\textit{glucose}~=~$[110-149]$). Moreover, this model is expected to produce a high percentage of false negatives (FN) for the lower and central values of glucose, while the risk of false positives (FP) is higher for higher values of the variable. Hence, the information provided by this EDP is highly valuable for an end-user accessing whether the model is truly adequate to be used, considering the distribution of FP and FN versus true positives\footnote{As a small side note we should remark that when applying the distribution rules to this same data set and model, we have obtained a set of rules that match the conclusions of this figure.}. In effect, for some applications FNs may be more costly than FPs, and vice versa, so being able to uncover these differences in performance as a function of the values of a predictor may be very useful for the end user. 

\subsection{Multivariate Methods}\label{sec:evalARs}

In this section we present a series of examples on using our distribution rules to discover and describe regions of the predictors space where the performance of a model is different, in a statistically significant way, from its global performance. Several types of differences can occur. Difference may consist of an area where the model is significantly  worse or better  than globally. It can also be sustained by cases where similar scores occur but with different error profile, i.e different confusion matrices. We will dwell into details on different aspects of local performance like false positive and false negative rates, precision and recall. An initial attempt to suggest procedures for model comparison in specific regions is also described.

Datasets with multiple class values frequently yield interesting rules. For instance, in dataset \textit{Glass} (c.f. Table \ref{tab:datasets}) that has 7 classes, for model NN,  the full dataset performance is represented by the following distribution:
\begin{verbatim}

CM={ 0/0.706,12/0.070,21/0.098,25/0.005,26/0.009,31/0.061,
32/0.014,52/0.005,57/0.005,62/0.005,65/0.005,
71/0.005,72/0.005,75/0.005,76/0.005 } 
\end{verbatim}

\noindent An example of a  distribution rule found by our methods is:

\begin{verbatim}
Ant sup = 0.03738  pvalue = 0.0187437297122432820000
CM={ 0/0.625,31/0.125,71/0.125,72/0.125 }    <--    
                      Al=[0.29 : 0.75]  &  K=[0.00 : 0.07]
\end{verbatim}
Here a local accuracy of 62.5\% (70.6\% in the full dataset) is observed for the subgroup described in the antecedent along with the following distribution of misclassifications: 31 for class 3 (12.5\%) and 71 and 72 for class 7 (also 12.5\%) each. Clearly the  error distribution of the subgroup described by this rule  is different  from the full data set performance distribution. This subgroup is composed of 8 test cases. Notice that attributes discretization is the same as described in Section \ref{sec:EDPs}. 


Continuing in the same dataset, some additional interesting subgroups were found. They also express a far different behaviour from the  general performance. The following rule represents one of these subgroups where we observe a prevalence of errors on class 3 that amount to  25\% of the errors in the subgroup (6.1\% in the full dataset). Also,   class 1 represents 16.7\% of the errors (7\% in the full dataset).

\begin{verbatim}
Ant sup = 0.05607  pvalue = 0.0275429788525971970000  
CM={ 0/0.333,12/0.167,21/0.083,31/0.250,52/0.083,57/0.083 }  <--    
                      Ca=[8.44 : 9.08]  &  Na=[13.08 : 13.72]  &  
                      Al=[1.25 : 1.66]

\end{verbatim}

On the other hand, some subgroups exhibit a unique type of error. An example is described by this rule:

\begin{verbatim}
Ant sup = 0.02804  pvalue = 0.0155661147914020260000 
CM={ 0/0.667,32/0.333 }    <--    
                      Na=[12.61 : 13.05]  &  K=[0.35 : 0.59]  &
                      Si=[72.97 : 73.44]  &  Ca=[8.44 : 9.08]

\end{verbatim}
where class 3 cases are often confused with class 2. There are other rules with a similar pattern for different classes, like the following one  for class 7 :

\begin{verbatim}
Ant sup = 0.01869  pvalue = 0.0197523002718819550000  
CM={ 0/0.750,76/0.250 }    <--    
                      K=[1.10 : 6.21]  &  Na=[13.73 : 14.46]
    
\end{verbatim}

\noindent or this rule for class 6:
\begin{verbatim}
Ant sup = 0.01869  pvalue = 0.0197523002718819550000  
CM={ 0/0.750,62/0.250 }    <--    
                    Ca=[11.32 : 16.19]  &  RI=[1.518 : 1.522]  &  
                    Al=[1.25 : 1.66]
                      
\end{verbatim}

All these rules find areas of the input space where some model behaves particularly bad for a certain class, which may be very useful for some applications.

\subsubsection{Model comparison. }

It is also very relevant to observe that on the same dataset different classification algorithms may show rather different behaviour. For instance, with a Random Forest on the previous dataset, we discover an unusually bad performance on class 2:

\begin{verbatim}
Ant sup = 0.01869  pvalue = 0.0010041075378298939000  
CM={ 0/0.250,21/0.750 }    <--    
                    RI=[1.511 : 1.516]  &  Na=[13.08 : 13.72]  &  
                    Ca=[7.90 : 8.43]
                      
\end{verbatim}

Again on the same dataset, this time for model NB, the following rule shows that no accurate predictions exists in this area of the predictors space.

\begin{verbatim}
Ant sup = 0.02336  pvalue = 0.0010024037427942200000  
CM={ 21/0.200,25/0.200,26/0.400,52/0.200 }    <--    
                       Ca=[11.32 : 16.19]  &  K=[0.08 : 0.33]  &  
                       RI=[1.522 : 1.534]
\end{verbatim}

For some application domains it may be very important to be able to  identify regions of the predictors space where specific types of errors occur. For instance, the following rule represents a subgroup in dataset \textit{Contraceptive} and model SVM where all errors involve mistakenly predicting the cases as class 1.

\begin{verbatim}

Ant sup = 0.01088  pvalue = 0.0010272093636637186000  
CM={ 0/0.438,21/0.125,31/0.438 }    <--    
                      a8=1  &  a2=3  &  a3=3
\end{verbatim}

\noindent This is rather different from the  global performance behaviour which is the following:

\begin{verbatim}

CM={ 0/0.556,12/0.030,13/0.122,21/0.060,23/0.090,31/0.095,
     32/0.048 }

\end{verbatim}






\subsubsection{Preconceptions}

When attributes have a clear meaning (like gender, age, race, etc) one can identify some interesting models' characteristics. As an example, rules can show model preconceptions. In dataset \textit{Adult}, the global performance of model NN is:

\begin{verbatim}
CM={ 0/0.851,12/0.048,21/0.100 }
\end{verbatim}

\noindent However, the  following rule stands out
\begin{verbatim}
Ant sup = 0.01268  pvalue = 0.0014509893634289416000  
CM={ 0/0.726,12/0.145,21/0.128 }    <--    
                      education=Doctorate

\end{verbatim}
showing model NN tending to frequently make the mistake of predicting high income (class 2) for people with PhDs. This subgroup is composed of 413 cases.

\begin{figure}[h!]
\begin{center}
  \includegraphics[width=0.99 \textwidth]{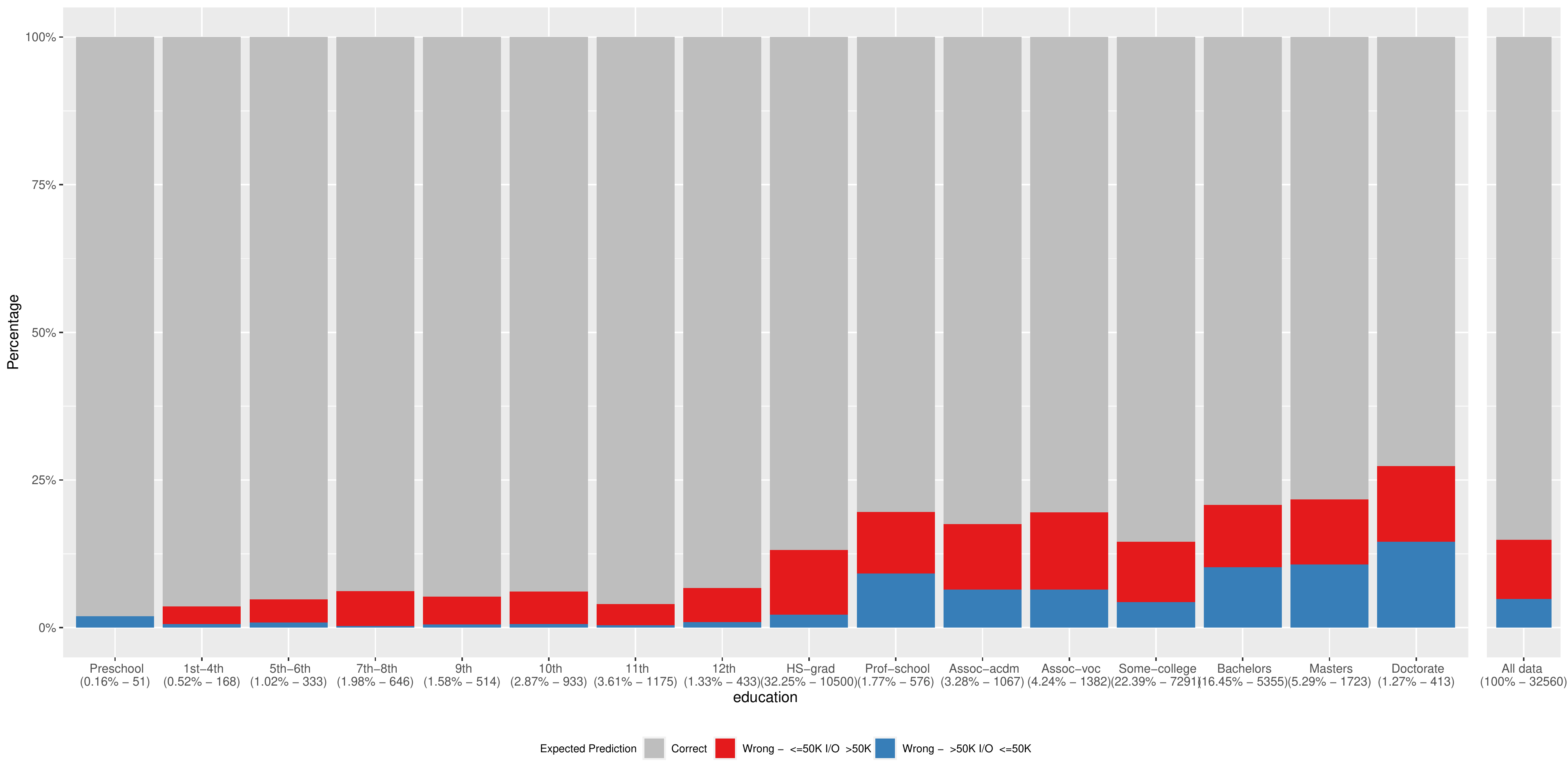}
\caption[]{Error Dependence Plot for feature \textit{education} from dataset \textit{Adult} trained with Neural Network.}
\label{fig:edp_adult}
\end{center}
\end{figure}

Figure \ref{fig:edp_adult}, depicting the EDP for the same predictor (feature \textit{education} from the NN model of dataset \textit{Adult}), corroborates visually this finding. In fact, this EDP shows that the margin of error is higher than the expected globally for any level of education higher than \textit{"HS-grad"} \footnote{Preschool $<$ 1st-4th $<$ 5th-6th $<$ 7th-8th $<$ 9th $<$ 10th $<$ 11th $<$ 12th $<$ HS-grad $<$ Prof-school $<$ Assoc-acdm $<$ Assoc-voc $<$ Some-college $<$ Bachelors $<$ Masters $<$ Doctorate}, and that overestimation (wrongly classifying a salary above of 50K) is most common for the cases with higher education (\textit{education} = Bachelors, \textit{education} = Masters and \textit{education} = Doctorate).

However, when the subgroup of male PhDs is considered this bias increases. That is,  the rate of type 12 error is higher (income is $\leq 50K$ but model predicts $>50K$).

\begin{verbatim}

Ant sup = 0.01004  pvalue = 0.0013500986880358072000  
CM={ 0/0.755,12/0.150,21/0.095 }    <--    
                      education=Doctorate  &  sex=Male
\end{verbatim}
\noindent This rule is supported by 326 cases.

\vspace{2mm}
Still within NN, young women income seem to be better captured  by the model when compared to all women. The improvement is significant.
\begin{verbatim}
    
Ant sup = 0.07703  pvalue = 0.0032418986092357450000  
CM={ 0/0.992,12/0.005,21/0.003 }    <--    
                      age=[17 : 24]  &  sex=Female

\end{verbatim}

\noindent versus the case of all women,
\begin{verbatim}
    
Ant sup = 0.33080  pvalue = 0.0032922890058757130000  
CM={ 0/0.926,12/0.022,21/0.052 }    <--    
                      sex=Female
\end{verbatim}

\vspace{2mm}
Continuing in the \textit{Adult} dataset and model NN, high number of education years and still in a productive age seems to lead to model confusion in both classes (with a significant number of errors). This subgroup is composed of 466 cases:
\begin{verbatim}

Ant sup = 0.01431  pvalue = 0.0011493800977763730000  
CM={ 0/0.783,12/0.109,21/0.107 }    <--    
                      education.num=[15 : 16]  &  age=[43 : 63]
\end{verbatim}

However, being a male helps to recover class 2. That is, the model tends to see this subgroup as low income people (class $\leq 50K$) but improves this misconception when the new condition is incorporated:

\begin{verbatim}
Ant sup = 0.01216  pvalue = 0.0011280584898949934000  
CM={ 0/0.811,12/0.111,21/0.078 }    <--    
                    education.num=[15 : 16]  &  age=[43 : 63]  &  
                    sex=Male
    
\end{verbatim}
On the other hand, changing the age range to younger people yields a higher bias on class 1 ($\leq 50K$):
\begin{verbatim}
    
Ant sup = 0.01314  pvalue = 0.0011326144060501797000  
CM={ 0/0.766,12/0.105,21/0.129 }    <--    
                      education.num=[15 : 16]  &  age=[25 : 42]
\end{verbatim}

In the same dataset, we analyse a race issue on income. Being white in a subgroup of married women arises difficulties for the model and yields slightly higher number of errors for both classes of income:
\vspace{2mm}

\begin{verbatim}
    
Ant sup = 0.04220  pvalue = 0.0023061284326488550000  
CM={ 0/0.710,12/0.122,21/0.168 }    <--    
                      sex=Female  &  
                      marital.status=Married-civ-spouse  &  
                      race=White
\end{verbatim}

\noindent against

\begin{verbatim}
    
Ant sup = 0.05089  pvalue = 0.0023578427029023910000  
CM={ 0/0.719,12/0.117,21/0.164 }    <--    
                      sex=Female  &  
                      marital.status=Married-civ-spouse
\end{verbatim}
However, model RF seems to be more robust in this white race situation:

\begin{verbatim}

Ant sup = 0.04220  pvalue = 0.0019779296158063914000 
CM={ 0/0.750,12/0.119,21/0.130 }    <--    
                      sex=Female  &  
                      marital.status=Married-civ-spouse  &  
                      race=White

\end{verbatim}

\subsubsection{Extreme performances}

Some regions of the predictors space lead to extreme performance situations. Below we show  several examples including both optimal or worst case scenario  performance.

One of such extreme situations is when we find sub-regions where a model does not make a single accurate prediction. For instance, in \textit{breast-base} with model SVM the following rules were derived:

\begin{verbatim}
    
Ant sup = 0.03750  pvalue = 0.0010004171846829683000  
CM={ 12/0.333,21/0.667 }    <--    breast.quad='right_up'  &  
                                   menopause='ge40'
Ant sup = 0.03750  pvalue = 0.0010004171846829683000  
CM={ 12/0.333,21/0.667 }    <--    age='60-69'  &  
                                   breast='right'  &  
                                   breast.quad='left_low'
Ant sup = 0.03750  pvalue = 0.0010004171846829683000  
CM={ 12/0.333,21/0.667 }    <--    tumor.size='25-29'  &  
                                   deg.malig='3'  & 
                                   breast.quad='left_low'
Ant sup = 0.03750  pvalue = 0.0010004171846829683000  
CM={ 12/0.333,21/0.667 }    <--    breast='right'  &  
                                   menopause='ge40'  &
                                   breast.quad='left_low'
Ant sup = 0.03750  pvalue = 0.0010004171846829683000  
CM={ 12/0.333,21/0.667 }    <--    tumor.size='25-29'  &  
                                   deg.malig='3'  &
                                   menopause='ge40'  &  
                                   node.caps='no'

\end{verbatim}

\noindent where classe 1 = ''recurrence-events''   and   classe 2 = ''no-recurrence-events''. Notice similar performance along quite different subgroups. The size of all these subgroups is 3 cases.

In \textit{BreastCancer} all models yield this rule (perfect 100\% hits):
\begin{verbatim}
    
Ant sup = 0.51775  pvalue = 0.0157340856655403500000  
CM={ 0/1.000 }    <--    Bare.nuclei=1  &  Normal.nucleoli=1
\end{verbatim}

\noindent the size of the subgroup is 350 cases. It is interesting to compare this rule with the EDP in Figure~\ref{fig:edp_breast}. In that plot we have observed that for $Normal.nucleoli=1$ there were serious and differentiated errors occurring. However, from this rule we observe that if on top of that characteristic the test cases show $Bare.nuclei=1$ then the models have no problems. This is an example of the complementarity of the two proposed analysis methods.

A similar example occurs on  dataset \textit{Fertility}, with  model SVM:

\begin{verbatim}

Ant sup = 0.30000  pvalue = 0.0473562617304469000000  
CM={ 0/1.000 }    <--    age=[0.56 : 0.64]
\end{verbatim}

\vspace{2mm}

Finally we show below a set of rules for the dataset \textit{breast-base} where the NN commits the same type of error on all cases:
\begin{verbatim}

Ant sup = 0.02500  pvalue = 0.0154383116883116850000  
CM={ 21/1.000 }    <--    breast.quad='right_up'  &  age='60-69'
Ant sup = 0.02500  pvalue = 0.0154383116883116850000  
CM={ 21/1.000 }    <--    breast.quad='right_up'  &  
                          menopause='ge40'  &
                          breast='right'
Ant sup = 0.02500  pvalue = 0.0154383116883116850000  
CM={ 21/1.000 }    <--    age='30-39'  &  deg.malig='3'  &  
                          breast='left'
Ant sup = 0.02500  pvalue = 0.0154383116883116850000  
CM={ 21/1.000 }    <--    age='30-39'  &  deg.malig='3'  &
                          menopause='premeno'
Ant sup = 0.02500  pvalue = 0.0154383116883116850000  
CM={ 21/1.000 }    <--    age='60-69'  &  deg.malig='2'  &
                          breast='right'
Ant sup = 0.02500  pvalue = 0.0154383116883116850000  
CM={ 21/1.000 }    <--    tumor.size='30-34'  &  age='50-59'  &
                          inv.nodes='0-2'
                          
Ant sup = 0.02500  pvalue = 0.0154383116883116850000  
CM={ 21/1.000 }    <--    irradiat='yes'  &  age='50-59'  &
                          inv.nodes='0-2'
Ant sup = 0.02500  pvalue = 0.0154383116883116850000  
CM={ 21/1.000 }    <--    irradiat='yes'  &  age='50-59'  &
                          node.caps='no'  &  menopause='premeno'

\end{verbatim}

\subsection{Discussion}

In this section we have presented several illustrations of the usefulness of the tools we have proposed for drilling down the analysis of the performance of black box classification models.

With classification EDPs we have shown that we are able to analyse the impact the domain of a single variable may have on the models' performance. This allows, for instance, the end user to anticipate critical situations if facing a test case with values on this predictor that are associated with poor performance.

Using distribution rules we are able to extend our deeper analysis of the performance of the models into a multivariate setting. Our proposed rules are able to uncover regions of the predictors' space where the performance of the models has high probability of being different from the global perspective given by standard  evaluation procedures. Compared to classification EDPs these rules allow us to explore interactions between multiple predictors that lead to differentiated performance. 

We see these proposed methods as decision support tools that end users can use to guide their decision on whether black box ML classification models can be trusted to drive their decisions. Contrary to global performance evaluation tools, our proposals drill down to specific test cases, which can be of key importance for application domains where accountability is essential.

\section{Conclusions}\label{sec:conclusion}

Reliable evaluation is a key step in any machine learning or data science project.  Being able to provide the end users with reliable estimates of the performance of the models is essential for the credibility of data analysts and of our research discipline. Nevertheless, end users also want to know why. Why is a model predicting a certain value? Why has the model made a mistake on this situation? As ML models are becoming more widely used, and as their complexity increases, these issues have become even more critical, particularly in application domains where they drive important and potentially costly decisions.

This paper presents two novel techniques that help in better assessing the reliability of the models even if they are black boxes and thus hard for humans to understand what drives their predictions. We specifically address the question: "what can I expect in terms of accuracy from my model given a test case with these properties?". Till now expected performance was assessed globally without any relationship to the predictor's values. We proposed two new techniques that drill down the performance analysis to understand how it depends on concrete predictor's values. Our illustrative cases studies with several datasets and ML models clearly show how these methods can help in uncovering and explaining unexpected behaviours of the models for some areas of the predictors' space. 

Our work has focused on blackbox classification models. However, this work can be easily extended to regression models. Moreover, they can obviously also be applied to models that are not blackboxes.

In the future we plan to extend our approaches to other predictive tasks (e.g. regression). Moreover, we think that our methods can be of use in a kind of case specific model selection strategy, where they can suggest that for a certain test case model A is to be preferred over model B, even-though model B is globally better.

\bibliographystyle{spmpsci.bst} 
\bibliography{main.bib}   

\end{document}